\documentclass[11pt]{article}

\usepackage{acl}

\usepackage{libertine} 

\usepackage{enumitem}
\usepackage{times}
\usepackage{latexsym}
\usepackage{graphicx}
\usepackage{subcaption}
\usepackage{calc}
\usepackage{hyperref}
\usepackage{roboto}
\usepackage[T1]{fontenc}

\usepackage[utf8]{inputenc}

\usepackage{microtype}

\usepackage{inconsolata}
\usepackage{adjustbox}
\usepackage{amsfonts}
\usepackage{graphicx}
\usepackage[most]{tcolorbox}
\usepackage{etaremune}

\usepackage[table]{xcolor}   
\usepackage{booktabs}       
\usepackage{multirow}       
\usepackage{array}
\usepackage{tikz}           
\usepackage{amsmath}
\renewcommand{\arraystretch}{1.2}

\definecolor{fullblue}{RGB}{255,255,255}
\definecolor{fullorange}{RGB}{255,180,140}

\usepackage{tikz} 
\usepackage[table]{xcolor}   

\newcommand{\HeatCell}[2]{%
  \pgfmathsetmacro{\rawval}{#1}%
  \pgfmathsetmacro{\clamped}{min(max(\rawval,0),50)}%
  \pgfmathsetmacro{\mixpct}{2*\clamped}%
  \edef\temp{\noexpand\cellcolor{fullorange!\mixpct!fullblue}}%
  \temp \libertineSB{#2}%
}

\newcommand{\diaghead}[1]{%
  \rotatebox[origin=c]{30}{\makebox[0pt][l]{#1}}%
}

\usepackage{tabularx}
\tcbuselibrary{skins}
\usepackage{ragged2e} 
\usepackage{caption}  
\definecolor{gtcol}{RGB}{10,72,143}
\definecolor{predcol}{RGB}{179,40,33}

\title{SemCSE-Multi: Multifaceted and Decodable Embeddings for Aspect-Specific and Interpretable Scientific Domain Mapping}

\author{Marc Brinner \\
  Computational Linguistics \\
  Bielefeld University, Germany \\
  {\tt marc.brinner@uni-bielefeld.de} \\\And
  Sina Zarrieß \\
  Computational Linguistics \\
  Bielefeld University, Germany \\
  {\tt sina.zarriess@uni-bielefeld.de} \\}
  \author{Marc Brinner \and Sina Zarrieß \\
  Computational Linguistics, Department of Linguistics\\
  Bielefeld University, Germany\\
  \texttt{\{marc.brinner,sina.zarriess\}@uni-bielefeld.de}}

\begin{document}
\maketitle
\begin{abstract}

We propose SemCSE-Multi, a novel unsupervised framework for generating multifaceted embeddings of scientific abstracts, evaluated in the domains of invasion biology and medicine. These embeddings capture distinct, individually specifiable aspects in isolation, thus enabling fine-grained and controllable similarity assessments as well as adaptive, user-driven visualizations of scientific domains. Our approach relies on an unsupervised procedure that produces aspect-specific summarizing sentences and trains embedding models to map semantically related summaries to nearby positions in the embedding space. We then distill these aspect-specific embedding capabilities into a unified embedding model that directly predicts multiple aspect embeddings from a scientific abstract in a single, efficient forward pass. In addition, we introduce an embedding decoding pipeline that decodes embeddings back into natural language descriptions of their associated aspects. Notably, we show that this decoding remains effective even for unoccupied regions in low-dimensional visualizations, thus offering vastly improved interpretability in user-centric settings.

\end{abstract}

\section{Introduction}
\label{sec:1}

The rapidly increasing volume of scientific publications \cite{bornmann_growth_2021} poses significant challenges for researchers in identifying relevant literature, gaining an overview of research trends, and navigating the diverse directions within a field.

Embedding models have been identified as a promising remedy: by mapping scientific abstracts into dense vector representations, these models enable efficient similarity assessment and support large-scale search as well as semantic clustering and visualization of a scientific domain \cite{cohan_specter_2020, ostendorff_neighborhood_2022, singh_scirepeval_2023, brinner2025semcse}.

However, current embedding approaches face two fundamental limitations, with the first one being the inherent lack of interpretability \cite{opitz_interpretable_2025}, making it difficult to discern the meaning associated with specific regions of the space.

\begin{figure*}[ht!]
    \centering
    \includegraphics[width=\textwidth]{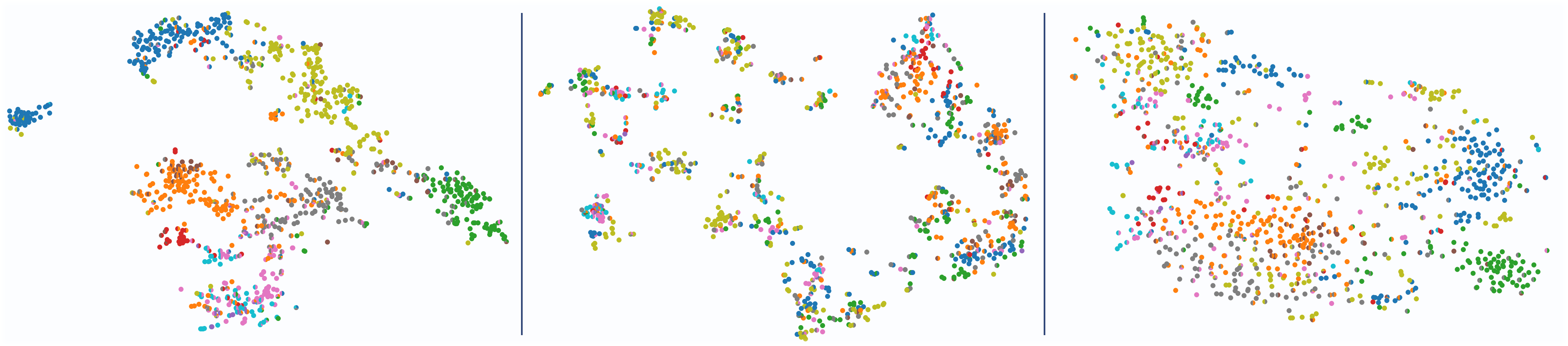}\vspace{2px}
    
    \begin{minipage}{0.33\textwidth}
        \centering
        Aspect: Hypothesis
    \end{minipage}%
    \begin{minipage}{0.33\textwidth}
        \centering
        Aspect: Species
    \end{minipage}%
    \begin{minipage}{0.33\textwidth}
        \centering
        SciNCL
    \end{minipage}

    \vspace{-3pt} 

    \caption{t-SNE visualizations of the embedding spaces of the unified embedding model for the \textit{hypothesis} and \textit{species} aspects, as well as for the SciNCL model \cite{ostendorff_neighborhood_2022}. Each dot represents a scientific study from the test set, with colors indicating the corresponding human-annotated hypothesis labels.}
    \label{fig:embedding_plots}\vspace{-10pt}
\end{figure*}

Second, they rely on an implicit, imprecise and fixed notion of similarity. Prior work has emphasized that similarity is inherently ill-defined unless the aspect under which it is evaluated is made explicit \cite{Goodman1972-GOOSSO, bar-etal-2011-reflective}. Nevertheless, existing models collapse the diversity of the many aspects into a single embedding, where abstracts are mapped to similar locations if, for instance, they cite each other \cite{cohan_specter_2020}, share authorship \cite{singh_scirepeval_2023}, or exhibit general (but underspecified) semantic relatedness \cite{brinner2025semcse}. As a result, 1) it is often unclear which factors are represented within two highly similar embeddings, and 2) it is not straight-forward to isolate specific aspects for targeted similarity assessments.

In this work, we mainly consider the scientific field of invasion biology, which investigates the introduction of non-native species into new ecosystems and their ecological, economic, and societal impacts. \textit{Hypotheses} about general causal relationships (e.g., the influence of ecosystem properties on the likelihood of species invasions) play a central role in this domain, so that a researcher might regard studies as related if they address the same hypothesis, regardless of other factors like the particular species or ecosystem under consideration.

Another researcher, however, may instead be interested in specific classes of \textit{species} or \textit{ecosystems}, without any requirements on the underlying hypothesis, thus regarding vastly different studies as highly related. Beyond these primary factors, additional aspects such as the precise \textit{research question}, specific \textit{recommendations}, or the \textit{methodology} employed can serve as important auxiliary indicators of similarity, helping to shape a more fine-grained and meaningful structure of the embedding space.

For this reason, a broadly useful embedding space that meets a user's requirements for literature retrieval or semantic visualization (see Figure \ref{fig:embedding_plots}) must support aspect-specific and user-adaptable similarity assessments, rather than relying on a single, pre-defined notion of similarity.

To address both adaptability and interpretability, we therefore propose a comprehensive framework for training a multifaceted embedding model that encodes distinct, user-specified aspects of scientific papers into separate embeddings, thus enabling precise, aspect-specific similarity assessment. Further, the framework supports the reconstruction of natural language descriptions from embeddings, thus providing direct interpretability of the embedding space and leading to controllable, interpretable, and interactive visualizations of scientific domains.

At a high level, our framework consists of three core components. First, building on recent advances in training semantically focused embedding models via LLM-generated summaries ({SemCSE}, \citealp{brinner2025semcse}), we train multiple \textbf{aspect-specific embedding models}. Each model focuses solely on one distinct aspect, and is trained to place multiple LLM-generated, aspect-specific summaries for a scientific abstract in close proximity within the embedding space (Section \ref{sec:3}).

Second, we distill these aspect-specific models into a \textbf{unified embedding model} called SemCSE-Multi, thus enabling a single model to jointly predict all aspect embeddings from a scientific abstract in one forward pass (Section \ref{sec:4}).

Third, we design a \textbf{decoding mechanism} that reconstructs natural language representations of given embeddings, thus substantially enhancing interpretability of the embedding space (Section \ref{sec:5}). Section \ref{sec:6} further extends this interpretability to low-dimensional visualizations, allowing for the decoding of previously unoccupied regions of the visualization into meaningful textual descriptions.

We evaluate our framework primarily in the domain of invasion biology, for which we had guidance from domain experts. To highlight its broader applicability, we further present a smaller-scale evaluation in the medical domain in Appendix \ref{sec:medical_appendix}.

\section{Related Work}

\textbf{Visualizing Scientific Domains} has been approached through a variety of methods. Citation networks treat papers that cite one another as related and cluster them accordingly \cite{van_eck_software_2010, eck_citnetexplorer_2014}, and various embedding models build on this idea by being trained on citation information to place papers that cite each other in close proximity within an embedding space, thereby enabling similarity comparisons without requiring direct access to the citation network \cite{cohan_specter_2020, ostendorff_neighborhood_2022, singh_scirepeval_2023}. However, relying on citation information as a proxy for similarity provides no insight or control into which aspects of the papers are being considered for assessing similarity.  

\textbf{Multifaceted Embeddings} have been proposed as a way to move beyond the rather simplistic definition of similarity in general embedding approaches. Existing approaches range from letting the model automatically determine relevant aspects to optimize downstream performance \cite{zhang_multi-view_2022, mysore_multi-vector_2022}, using the location of a citation within a paper as an indicator of the aspect being referenced \cite{10.1007/978-3-319-31750-2_42,ostendorff_aspect-based_2020}, creating separate embeddings for different sections of a paper \cite{singh_cosaemb_2024}, or relying on supervised datasets to train embedding models specialized for particular aspects or task types \cite{jain_learning_2018, risch_multifaceted_2021, ostendorff_specialized_2022, singh_scirepeval_2023}.  

A common limitation across both general and multifaceted embedding models is the lack of user control in specifying which aspects of similarity should be taken into account, which is only possible within methods that rely on large supervised datasets. Our work addresses this limitation by introducing a fully unsupervised method for constructing multifaceted embeddings that explicitly encode the aspects most relevant to potential users.

Lastly, \textbf{Decoding Embeddings} back into natural language is crucial for enhancing the interpretability of our proposed embedding framework. While recent work has explored embedding inversion primarily in the context of privacy risks \cite{li_sentence_2023, morris_text_2023, huang_transferable_2024}, its potential for improving interpretability has remained largely unexplored.

\section{Training Aspect Embedding Models}
\label{sec:3}

The first step in our embedding framework is the creation of \textit{several distinct} embedding models, each capable of mapping summaries of specific aspects into a dedicated semantic embedding space. The six aspects under consideration were motivated in Section \ref{sec:1} and are detailed in Appendix \ref{app:aspects}.

\subsection{Dataset Creation}

The foundation of our framework is a dataset of scientific abstracts from the target domain. For this purpose, we leverage a corpus of 37,786 papers from the field of invasion biology, compiled by \cite{brinner2025enhancing}. To train the aspect-specific embedding models, we then require multiple summarizing sentences for each abstract and aspect, that summarize the aspect-specific information of that paper for \textit{just the aspect of interest in isolation}, while excluding information about other factors.

To generate these summaries, we use Mistral 3.1 Small \cite{mistral}, prompting it four times per abstract to summarize the designated aspect or reply “Not applicable” if the aspect does not apply to the study. Due to stochasticity in autoregressive generation, the resulting summaries differ slightly in phrasing but remain highly semantically aligned because of the shared generation context. Important insights, exemplary prompts and dataset sizes for each factor are reported in Appendix \ref{app:inv_data_creation}, with exemplary summaries being displayed in Figure \ref{fig:invasion_bio_example}.

\subsection{Model Training}

\label{sec:individual_model_evaluation_setup}

\begin{table*}[ht!]
\renewcommand{\arraystretch}{0.90}
\centering
\resizebox{\textwidth}{!}{
\begin{tabular}{lcccccc}
\toprule
 & Hypotheses & Ecosystem & Res. Question & Species & Methodology & Recommendation \\
\midrule
Samples & 954 & 951 & 954 & 954 & 954 & 63 \\
MRR \footnotesize Main model & .673 & .620 & .879 & .589 & .547 & .982 \\
MRR \footnotesize Other models & .556 & .404 & .757 & .374 & .352 & .903 \\
MRR \footnotesize SemCSE & .539 & .377 & .780 & .349 & .281 & .838 \\
MRR \footnotesize SciDeBERTa & .228 & .104 & .169 & .106 & .075 & .426 \\
\bottomrule
\end{tabular}}
\caption{Retrieval performance of the individual summary embedding models measured via Mean Reciprocal Rank. Main = model trained for the aspect, Others = average of non-matching models.}
\label{tab:individual_retrieval}\vspace{-10px}
\end{table*}

For each aspect separately, we train a dedicated embedding model following the SemCSE approach of \cite{brinner2025semcse}. Specifically, we apply a contrastive loss that encourages the model to place two aspect summaries from the same abstract close together in the embedding space, while pushing apart embeddings of unrelated summaries (i.e., same aspect but different abstract):\vspace{-10px}

\begin{align*}
    \mathcal{L} = - \frac{1}{|\mathcal{B}|}\sum_{i \in \mathcal{B}} \log \frac{{\textrm e}^{\text{sim}(e_i, e_i^+) / \tau}}{\sum_{j} {\textrm e}^{\text{sim}(e_i, e_j^+) / \tau}}
\end{align*}

Here, the loss is computed over a batch $\mathcal{B}$ consisting of $|\mathcal{B}|$ pairs of related samples with embeddings $(e_i, e_i^+)$. As the similarity function, we use cosine similarity, with $\tau$ being a temperature parameter. As the base encoder, we employ SciDeBERTa \cite{10355927}, mapping the \texttt{CLS} token embedding down to $150$ dimensions with an additional linear projection head. Additional implementation and training details are provided in Appendix \ref{app:aspect_training}.

\subsection{Evaluation Setup}

Since the aspect-specific embedding models are merely an intermediate result, we perform a small-scale evaluation in a retrieval setting. To this end, we are using a separate test set of 954 invasion biology abstracts \cite{brinner2022linking} with aspect-specific summaries  (generated as for training).

To evaluate each aspect, we then randomly select two summaries per abstract, with the first being used as a query and the second as a candidate in the retrieval task. All summaries are embedded using the corresponding aspect-specific model, and - for each query - the task is to retrieve the correct match from the complete pool of up to 954 candidates. Performance is measured using the mean reciprocal rank $\text{MRR} = \frac{1}{|Q|} \sum_{i=1}^{|Q|} \frac{1}{\text{rank}_i}$, where $|Q|$ denotes the number of queries and $\text{rank}_i$ is the position at which the correct candidate is retrieved for query $i$.

\subsection{Results}

Results for the retrieval-based evaluation are reported in Table \ref{tab:individual_retrieval}. Overall, the models achieve strong retrieval performance, with MRR values exceeding 0.5, indicating that the correct match is typically retrieved within the top two candidates.

To underscore the aspect-specific nature of the models, we conduct an ablation study in which, for each aspect, retrieval is performed using all models trained on the other aspects. In this setting, performance drops substantially, thus providing clear evidence that each model learns to capture information specific to its designated aspect. We also evaluate SemCSE as baseline, which was trained on general scientific texts and cannot match the aspect-specific models as well, and the base SciDeBERTa decoder, which achieves the lowest scores.

While these results are promising, it is important to note that the models at this stage operate exclusively on summarizing sentences, which does not yet reflect typical downstream applications. Their primary role is therefore to shape a semantic embedding space that encodes domain- and aspect-relevant features, thereby enabling high-quality assessment of aspect-specific similarity.

\section{Distilling a Unified Embedding Model}
\label{sec:4}

In the previous section, we introduced several distinct aspect-specific embedding models, each trained to embed summarizing sentences related to a single aspect. In this section, we distill the aspect-specific embedding capabilities into a single \textit{unified embedding model} called SemCSE-Multi that directly predicts aspect-specific embeddings from the full abstract, thereby eliminating the costly and inconsistent dependence on generated summaries.

\subsection{Model Training}

Training the unified model is based on the same dataset of 37,786 abstracts with corresponding aspect-specific summaries. For each abstract and aspect, we compute a single aspect-specific target embedding by averaging the embeddings of the individual summaries produced by the corresponding aspect-specific model. SemCSE-Multi is then trained to reconstruct these aspect-specific embeddings directly from the scientific abstract.

Concretely, we employ a pretrained SciDeBERTa encoder \cite{10355927} with several linear projection layers - one for each aspect - that map the 768-dimensional \texttt{CLS} token embedding into the 150-dimensional aspect-specific embeddings. The model is trained using an L2 loss between the predicted and target embeddings. Further experimental details are provided in Appendix \ref{app:unified_training}.

\subsection{Evaluation Setup}
\label{sec:unified_evaluation_setup}

\begin{table*}[ht]
\vspace{5px}
\centering
\libertine
\footnotesize
\setlength{\tabcolsep}{14pt} 
\renewcommand{\arraystretch}{1.25}
\libertineSB{
\begin{tabular}{clccccccc}
\multicolumn{2}{r}{\raisebox{0em}[0pt][0pt]{\makebox[0pt][r]{\shortstack[r]{Ground Truth\\Similarities $\rightarrow$}}}} \hspace{-25px} & \diaghead{Hypo-Labels} & \diaghead{Hypothesis} & \diaghead{Ecosystem} & \diaghead{Res. Question} & \diaghead{Species} & \diaghead{Methodology} & \diaghead{Recom.} \\
\midrule
\multirow{1}{*}{\hspace{-10px}\rotatebox{90}{} \hspace{-30px}}
  & \hspace{-20px}Hypo-Labels & \HeatCell{100}{100.0} & \HeatCell{35}{35.1} & \HeatCell{9}{8.7}  & \HeatCell{38}{37.8} & \HeatCell{10}{9.8}  & \HeatCell{19}{19.4} & \HeatCell{6}{5.8} \\ \cline{1-2}
  \multirow{6}{*}{\hspace{-10px}\rotatebox{90}{\underline{SemCSE-Multi}} \hspace{-30px}} & \hspace{-20px}Hypothesis & \HeatCell{42}{42.0} & \HeatCell{44}{44.0} & \HeatCell{21}{21.3} & \HeatCell{46}{45.5} & \HeatCell{31}{31.2} & \HeatCell{27}{26.6} & \HeatCell{8}{8.2}  \\
  & \hspace{-20px}Ecosystem & \HeatCell{14}{13.5} & \HeatCell{16}{15.7} & \HeatCell{37}{37.3} & \HeatCell{15}{14.6} & \HeatCell{23}{23.2} & \HeatCell{16}{16.1} & \HeatCell{4}{4.1}  \\
  & \hspace{-20px}Research Question & \HeatCell{35}{34.9} & \HeatCell{38}{38.2} & \HeatCell{26}{25.5} & \HeatCell{40}{40.0} & \HeatCell{29}{28.6} & \HeatCell{24}{24.0} & \HeatCell{8}{8.3}  \\
  & \hspace{-20px}Species & \HeatCell{15}{15.0} & \HeatCell{17}{16.9} & \HeatCell{31}{30.8} & \HeatCell{17}{16.5} & \HeatCell{45}{44.8} & \HeatCell{21}{20.7} & \HeatCell{4}{4.3}  \\
  & \hspace{-20px}Methodology & \HeatCell{19}{18.6} & \HeatCell{22}{21.8} & \HeatCell{11}{11.4} & \HeatCell{27}{27.1} & \HeatCell{14}{13.8} & \HeatCell{47}{46.7} & \HeatCell{9}{9.0}  \\
  & \hspace{-20px}Recommendation & \HeatCell{33}{32.7} & \HeatCell{36}{36.2} & \HeatCell{26}{26.1} & \HeatCell{37}{37.0} & \HeatCell{30}{30.4} & \HeatCell{28}{27.5} & \HeatCell{8}{7.8} \\
  \midrule
  & \hspace{-20px}SemCSE-Multi Combined & 42.0 & 44.0 & 37.3 & 40.0 & 44.8 & 46.7 & 7.8 \\ \midrule
  \multirow{5}{*}{\hspace{-10px}\rotatebox{90}{\underline{Baselines}} \hspace{-30px}}
  & \hspace{-20px}SPECTER & 18.9 & 26.1 & 31.5 & 28.9 & 34.5 & 24.0 & 5.1 \\
& \hspace{-20px}SPECTER2 & 22.1 & 27.4 & 26.3 & 30.1 & 30.4 & 19.5& 8.3\\
& \hspace{-20px}SciNCL & 20.7& 28.6& 30.4& 30.2& 37.5& 23.6& 6.3\\
& \hspace{-20px}SemCSE & 22.9& 35.0& 26.2& 37.1& 29.3& 31.9& 4.5\\
& \hspace{-20px}InvDef-DeBERTa & 21.0 & 32.2 & 26.8 & 37.8 & 29.7 & 30.9 & 9.4 \\
\bottomrule
\end{tabular}}\vspace{-3px}

\caption{Correlation matrix showing Spearman correlations between pairwise similarities computed from embedding models (aspect embeddings from SemCSE-Multi as well as baselines) and the pairwise ground-truth similarities, which are computed from the human-annotated hypothesis labels ("Hypo-Labels") or LLM-generated (all others).}
\label{tab:predictor_label_heatmap}\vspace{-8px}
\end{table*}

We evaluate the unified embedding model with respect to its ability to capture aspect-specific similarities between samples in the resulting embedding spaces. For this purpose, we make use of two complementary sources of target similarity:

\begin{enumerate}
    \setlength{\itemsep}{2pt}
    \setlength{\parskip}{2pt}
    \item The test set of 954 invasion biology abstracts is annotated by domain experts with labels indicating which of the ten most prominent hypotheses each abstract addresses. Thus, for the \textit{hypothesis} aspect, the model should place abstracts that address the same hypothesis in close proximity within the embedding space.
    \item For the remaining aspects, high-quality expert annotations are difficult to obtain at scale. We therefore rely on LLM-generated pairwise similarity assessments between all pairs of abstracts from a set of 250 samples.
\end{enumerate}

In the latter case, Qwen3-30B-A3B \cite{yang2025qwen3technicalreport} is prompted to assign integer similarity scores to pairs of abstracts, strictly with respect to a given aspect and based on precise category definitions (Appendix \ref{app:LLM_assessments}). While these assessments are less reliable than the expert annotations, strong correlations with embedding-based similarities nevertheless indicate that the embeddings capture aspect-specific information. Further, low alignment of embedding similarity for one aspect with LLM-based scores for a different aspect suggests that the SemCSE-Multi can effectively disentangle and isolate aspect-specific information.

For evaluation, we compute similarities between a given abstract and all others in the dataset using the aspect-specific embeddings, and compare the resulting similarity ranking to that induced by the ground-truth similarity labels (either human-annotated or LLM-generated) using Spearman correlation. For the hypothesis annotations, pairwise similarity is defined as 1 if two abstracts address the same hypothesis and 0 otherwise.

\subsection{Results}

Results are displayed in Table \ref{tab:predictor_label_heatmap}. The upper part reports the correlations between the aspect-specific embeddings created by SemCSE-Multi and the different ground-truth similarity assessments. For completeness, we include the human-annotated hypothesis labels as predictors (row 1), allowing us to assess their correlation with the LLM-produced similarity scores. We see that the human-annotated hypothesis labels show high correlation with the LLM-annotated \textit{hypothesis} similarity scores, thus demonstrating that the LLM-based annotation is a reasonable approach. 

The first key observation is that the highest correlation values consistently occur along the diagonal, clearly indicating that embeddings for a given aspect align most strongly with similarity assessments for that same aspect. A small number of notable off-diagonal correlations can be explained by genuine overlaps between factors:

\begin{enumerate}[itemsep=0.0em, topsep=0.2em, labelsep=0.2em]
    \item As discussed, the human-annotated hypothesis labels strongly correlate with the LLM-based similarities for the \textit{hypothesis} aspect.
    \item Embeddings for the \textit{hypothesis} aspect correlate well with the human-annotated hypothesis labels, thus demonstrating that this factor is effectively represented in the embeddings.
    \item A strong alignment is observed between the \textit{hypothesis} and \textit{research question} aspects, reflecting their natural interdependence: in most studies, the specific hypothesis under investigation is directly tied to the research question.
\end{enumerate}

The substantially lower correlations observed in the remaining off-diagonal cells provide strong and statistically significant (Appendix \ref{app:statistical_sig}) evidence that the aspect-specific embeddings are indeed isolating the intended factors. In consequence, this disentanglement enables controlled similarity assessments adhering to specific user needs.

One notable exception is the \textit{recommendation} aspect, where embeddings only show weak alignment with the LLM-based similarities. This is likely due to the scarcity of explicit recommendations in scientific abstracts, which reduces the number of training samples, and also reduces data quality due to the increased the likelihood of the LLM hallucinating summaries if recommendations are absent or solely hinted at within the abstract.


To contextualize the strength of the observed correlations, we additionally evaluate several widely used scientific embedding models: SPECTER \cite{cohan_specter_2020}, SPECTER2 Proximity \cite{singh_scirepeval_2023}, SciNCL \cite{ostendorff_neighborhood_2022}, SemCSE \cite{brinner2025semcse}, and the invasion biology-specific InvDef-DeBERTa \cite{brinner2025enhancing}. We do not include other multifaceted models, as (to our knowledge) none of them allow for freely selecting the target aspects without access to corresponding supervised datasets.

Across all aspects except \textit{recommendation}, none of these models achieve correlations comparable to our aspect-specific embeddings, thus demonstrating the benefits of having an aspect-specific focus. More importantly, these baseline models cannot disentangle specific aspects, making them unsuitable for tasks requiring fine-grained similarity assessments. This is underscored by Figure \ref{fig:embedding_plots}, visualizing the clearly differing and semantically structured aspect-specific embedding spaces by SemCSE-Multi for the \textit{Hypothesis} and \textit{Species} aspects, as well as the substantially less distinctly structured space induced by the SciNCL model. Together, these results provide clear evidence that a multifaceted embedding model offers distinct advantages both in terms of accuracy and in the ability to isolate aspect-specific information.

We further find that using a two-stage procedure is crucial for disentangling the individual aspects. We evaluated an additional baseline (Appendix \ref{app:baseline}) that directly trains a unified model for all aspects, but find that correlations between different aspects are substantially higher, likely caused by the co-adaption of the model to all aspects at once, which leaks information between them. Training aspect-specific models in isolation and distilling a unified model to predict these properly isolated embeddings is therefore a crucial advantage.

\section{Embedding Decoding}
\label{sec:5}

Embedding spaces are inherently abstract: it is often unclear what semantic meaning a small perturbation in the space represents, or why a longer text is assigned to a particular position, thus substantially limiting the potential for interactive exploration. Our framework, however, provides a unique opportunity to overcome this limitation: For each aspect-specific embedding in the training data, we have access to matching summarizing sentences that describe precisely that aspect, and therefore provide interpretable natural language descriptions of the information encoded in that embedding.

In this section, we demonstrate that it is possible to use a decoder LLM to map aspect-specific embeddings back into the space of natural language descriptions, thus substantially enhancing the interpretability of the embedding space.

\subsection{Model Training}

Our embedding decoding pipeline builds on the autoregressive language generation capabilities of LLMs. To enable decoding, we first map the content of a given aspect-specific embedding into the input embedding space of the LLM. This is achieved using a dense feed-forward network with a single hidden layer, which projects the input embedding into a sequence of five embedding tokens that are provided as input embeddings to the LLM.

To guide the decoding process, we additionally prepend five trainable, input-independent embeddings that function as general task descriptors, and optionally append one more token to act as delimiter to the decoded text, in case a normal autoregressive language model is used instead of a chat model. During training, only these embeddings and the mapping network are optimized, while the parameters of the underlying LLM remain frozen. Using the dataset of summarizing sentences paired with their corresponding aspect-specific embeddings, the model is trained with a standard language modeling cross-entropy loss to reconstruct the original summary from its embedding representation.

We evaluate two LLM backbones of different sizes and training paradigms, with one being a purely autoregressive language model in the form of Llama 3 8B \cite{grattafiori2024llama3herdmodels}, and the other being a larger (24B parameters) chat model in the form of Mistral Small 3.1 \cite{mistral}.

\subsection{Evaluation Setup}
\label{sec:decoder_eval_setup}

\begin{table*}[ht]
\setlength{\tabcolsep}{2pt} 
\centering
\renewcommand{\arraystretch}{0.95}
\resizebox{\textwidth}{!}{
\begin{tabular}{lccccccc}
\toprule
& \multicolumn{4}{c}{Llama} & \multicolumn{3}{c}{Mistral} \\
\cmidrule(lr){2-5} \cmidrule(lr){6-8}
Aspect & Matching & Reconstructed & Shuffled & Unconditioned & Matching & Reconstructed & Shuffled \\
\midrule
\multicolumn{8}{c}{\textbf{Perplexity}} \\ \hline
Hypotheses \small general & 5.86 & 8.16 & 22.52 & 53.88 & 5.43 & 8.07 & 25.21 \\
Hypotheses \small specific & 5.01 & 6.32 & 14.53 & 24.44 & 4.52 & 5.96 & 14.97 \\
Ecosystem & 5.39 & 7.12 & 16.50 & 46.97 & 5.38 & 7.13 & 17.05 \\
Res. Question & 5.76 & 8.67 & 21.80 & 27.18 & 5.93 & 8.43 & 20.62 \\
Species & 5.00 & 5.94 & 13.94 & 37.87 & 5.24 & 6.23 & 14.92 \\
Methodology & 5.37 & 6.67 & 16.37 & 94.08 & 5.58 & 6.64 & 16.94 \\
Recommendation & 13.14 & 17.82 & 27.14 & 65.50 & 12.15 & 16.72 & 27.05 \\ \midrule
\multicolumn{8}{c}{\textbf{BERTScore}} \\ \hline
Hypotheses \small general & 77.5 & 76.1 & 71.9 & - & 79.2 & 77.6 & 72.9 \\
Hypotheses \small specific & 76.5 & 75.5 & 71.6 & - & 78.0 & 76.5 & 71.2 \\
Ecosystem & 77.2 & 76.1 & 73.0 & - & 78.6 & 77.4 & 73.7 \\
Res. Question & 74.5 & 72.3 & 69.0 & - & 75.9 & 74.1 & 70.4 \\
Species & 75.5 & 74.5 & 68.8 & - & 76.8 & 75.5 & 69.7 \\
Methodology & 78.7 & 77.8 & 72.5 & - & 79.9 & 78.9 & 72.9 \\
Recommendation & 70.1 & 69.4 & 67.8 & - & 70.6 & 69.0 & 68.5 \\
\bottomrule
\end{tabular}}\vspace{-3px}
\caption{Perplexity and BERTScore results for Llama and Mistral decoders when decoding embeddings under normal, reconstructed and randomized settings. Lower perplexity indicates better performance, while higher BERTScores indicate better alignment of generated descriptions with ground-truth descriptions.}
\label{tab:decoder_perplexity}\vspace{-10px}
\end{table*}

We evaluate the decoding pipeline on the test set of 954 abstracts for which we predict aspect embeddings with SemCSE-Multi. For a given aspect embedding, we first measure the perplexity of the ground-truth summaries under the predicted distribution produced by the LLM, so that low perplexity indicates that the decoder can recover the semantic content of the original summary with high probability. As a second evaluation metric evaluating actual generated sentences, we use BERTScore \cite{DBLP:conf/iclr/ZhangKWWA20} on a pretrained SciBERT model \cite{beltagy-etal-2019-scibert}, which evaluates the correspondence between generated and ground-truth tokens.

To verify that the model does not merely generate generic aspect-related sentences that achieve reasonable performance while ignoring the actual embedding input, we additionally evaluate both metrics when matching the embedding of one sample with the summaries of another (\textit{Shuffled}), and, for the pure autoregressive Llama model, without any embedding information (\textit{Unconditioned}). If the trained decoding process indeed reconstructs the semantic meaning of the embeddings, using \textit{matching} embedding and ground-truth should result in substantially lower perplexity and higher BERTScores compared to the control experiments.

\subsection{Results}

The results for embedding decoding are reported in Table \ref{tab:decoder_perplexity}. We observe consistently low perplexity values when reconstructing the summarizing sentences from the \textit{matching} embeddings, demonstrating that the decoder is able to accurately recover the aspect-specific semantic information.

In contrast, both the \textit{Shuffled} and \textit{Unconditioned} control conditions yield markedly higher perplexity scores, thus clearly confirming that the decoder assigns a higher likelihood to the ground-truth reconstruction if conditioned on the correct embedding. The BERTScore results show a similar picture, with BERTScores in the \textit{matching} setting being consistently and substantially higher than those in the \textit{shuffled} setting. This demonstrates that not only the likelihood of the ground truth is increased when conditioned on the correct embedding, but that the actual generations show a good alignment with the original ground-truth. Together, these results establish that embedding decoding provides a reliable and interpretable link between aspect-specific vector representations and their corresponding semantic descriptions. Exemplary reconstructions of summarizing sentences are included in Figure \ref{fig:invasion_bio_example}.

\section{Interactive Embedding and Decoding}
\label{sec:6}

Embedding spaces are typically high-dimensional, which is essential for encoding a wide range of semantic factors across their dimensions. However, for user-centric applications, these spaces must be projected into lower-dimensional representations that allow for intuitive visualization while preserving, as faithfully as possible, the similarity structure of the original high-dimensional space.  

In this section, we demonstrate how dimensionality reduction can be combined with our decoding approach to generate semantic natural language explanations of potentially unoccupied regions within a visualization. To this end, we introduce three techniques that enable meaningful interaction with the resulting visualizations, laying the foundation for the subsequent evaluation.

\subsection{t-SNE}

To obtain low-dimensional visualizations of the embedding space, we use the t-SNE method \cite{vandermaaten08a}. The method begins by computing a pairwise distance matrix $A$ in the original high-dimensional space, which is converted into symmetric similarity probabilities $P$ that quantify how strongly two samples are related. It then optimizes low-dimensional coordinates $Y$ so that their corresponding similarity probabilities $Q$ mirror those from the high-dimensional space:\vspace{-10px}

\begin{align}
\label{eq:tsne_optim}
\mathcal{L}(Y) = \mathrm{KL}(P \,\|\, Q) = \sum_{i \neq j} p_{ij} \log \frac{p_{ij}}{q_{ij}}
\end{align}\vspace{-10px}

This objective ensures that points which are neighbors in the original embedding space remain close in the visualization, while unrelated points are pushed apart. For details, see Appendix \ref{app:tsne}.

\subsection{Methods for Interaction}

The fully differentiable and optimization-based nature of t-SNE provides several key capabilities that enable interactive exploration of the visualization.  

First, \textbf{weighting different aspects} is straightforward by adjusting the distance function used to compute the distance matrix $A$. Using individual, aspect-specific distances $d_a(x_i, x_j)$ between samples $i$ and $j$ for aspect $a$, we define a combined distance as $d(x_i, x_j) = \sum_a \alpha_a d_a(x_i, x_j)$ with $\sum_a \alpha_a = 1$, so that modifying the values of $\alpha$ effectively prioritizes certain aspects over others, producing a low-dimensional visualization that reflects these relative importances. Figure \ref{fig:embedding_plots} demonstrates the vastly differing visualizations that result from distinct aspect-specific weightings.


Second, \textbf{embedding additional user-provided samples} is done by extending the matrix $A$ with the distances of the new sample in the embedding space and optimizing Equation \ref{eq:tsne_optim} with respect to just the new low-dimensional coordinates.  

Finally, we can \textbf{reconstruct high-dimensional embeddings} for user-specified low-dimensional points in the visualization. To do this, all existing high-dimensional embeddings are kept fixed, and both an additional high-dimensional embedding $e_{n+1}$ as well as the low-dimensional user-selected point $y_{n+1}$ are added. Due to the differentiability of computing $A$ and $P$, the embedding $e_{n+1}$ can then be trained to minimize Equation \ref{eq:tsne_optim}, resulting in a high-dimensional embedding that would map to the user-specified point in the visualization.

Once obtained, this reconstructed embedding can be passed through the decoding pipeline to generate a natural language description, thereby revealing the semantic meaning associated with any location in the visualization. For technical details and practical considerations regarding the optimization, see Appendix \ref{app:interactive_details}.

\subsection{Evaluation Setup}
\label{sec:t_SNE_evaluation_setup}

We evaluate our framework’s ability to decode unoccupied positions in low-dimensional visualizations, thereby making previously unoccupied regions interpretable.

This is done by creating a t-SNE visualization for a single aspect (i.e., $\alpha_a = 1$ and all other weights are $0$) and mapping an additional sample represented by embedding $e_{n+1}$ into the visualization, yielding low-dimensional coordinates $y_{n+1}$. Afterwards, we remove that sample again and reconstruct a new high-dimensional embedding $\tilde{e}_{n+1}$ for the (now empty) position $y_{n+1}$. We then evaluate our decoder on embedding $\tilde{e}_{n+1}$, but taking aspect-specific summaries for embedding $e_{n+1}$ as targets.

In essence, this procedure therefore tests whether empty positions in the visualization, once decoded, yield semantic descriptions consistent with those of real samples that would occupy the same location. Further details are specified in Appendix \ref{app:reconstruc_experiment}.

\subsection{Results}

Evaluation results are presented in Table \ref{tab:decoder_perplexity}. As expected, the \textit{reconstructed} scores are slightly underperforming those obtained when decoding directly from the original (\textit{matching}) high-dimensional embeddings. This is unsurprising since the two-dimensional projection is not injective, so that multiple high-dimensional points could map to the same low-dimensional location. Nevertheless, both scores are approaching those of direct reconstruction and, most importantly, substantially outperform the \textit{shuffled} control setup, thus demonstrating that the proposed approach effectively generates accurate natural language representations for arbitrary positions within the embedding visualization.


\section{Conclusion}

In this work, we introduced a comprehensive framework for constructing multi-faceted embeddings of scientific texts. Our approach directly addresses the limitations of general-purpose embeddings, which fail to capture and isolate nuanced, aspect-specific similarities that are crucial for scientific analysis. By explicitly disentangling different aspects of a study, our framework enables user-controlled, aspect-weighted similarity scoring and supports visualizations that reflect these adaptive similarities.

A central advantage of our framework is the vastly improved interpretability of both high-dimensional embeddings and low-dimensional coordinates in visualizations, enabled through a decoding procedure producing accurate semantic descriptions of a given point. Crucially, this is solely made possible by the unique training setup of our framework, which provides a large dataset of embeddings with corresponding summarizing sentences for supervised decoder training.

Based on the extensive experiments in the domain of invasion biology, complemented by evaluations in the medical domain (Appendix \ref{sec:medical_appendix}), we believe that the improved capabilities for accurate and aspect-specific similarity assessments are applicable to a variety of domains, thus offering researchers powerful new tools to navigate and interpret large bodies of literature.

Further, while our study focused on scientific texts, the principles underlying the framework are not domain-specific, so that future work may extend these ideas to other research areas, adapt them to non-scientific domains, or even non-textual modalities such as images.

\section{Limitations}

We note several limitations of this work and of the proposed embedding framework.

First, although the work by \cite{brinner2025semcse} has demonstrated strong performance using LLM-generated summaries of scientific texts, and have argued that summary quality exerts only a minor influence on embedding quality, the use of such summaries nevertheless introduces a potential source of uncertainty. In particular, biases inherent to large language models may subtly affect the representation and alignment of information, thereby influencing the learned embeddings in ways that are difficult to quantify.

Second, our experiments in the domain of invasion biology revealed that aspects represented in only a small fraction of the training data can lead to degraded embedding quality. This effect is likely attributable both to the substantially reduced number of training samples for that aspect and to inaccuracies in the LLM-generated summaries, which is most prominent in cases where the aspect is not clearly addressed in the scientific abstract. In these cases, the LLM occasionally produces summaries that are not sufficiently supported by the original abstract, which can weaken the semantic alignment between summary pairs and thus reduce the quality of the aspect-specific embedding model trained on them.

Finally, our experiments on the medical domain highlight the critical importance of prompt design in summary generation. High-quality prompts are essential to ensure that the resulting embeddings not only match strongly similar samples (e.g., abstracts that address the exact same disease), but also accurately rank less related abstracts with respect to more subtle factors. The key to induce the desired behavior is the design of prompts that reliably encode all of these relevant aspects, which generally requires careful consideration and supervision by domain experts.

\bibliography{custom}

\appendix

\section{Experimental Details: Invasion Biology }
The precise code implementation with all technical details, prompts, hyperparameters, trained model checkpoints, etc. is available at 
\href{https://github.com/inas-argumentation/SemCSE-Multi}{github.com/inas-argumentation/SemCSE-Multi}.

\subsection{Aspects}
\label{app:aspects}

With the help of domain experts, we identified six aspects of invasion biology that capture complementary dimensions of research studies that are relevant for assessing similarities between them. These aspects were chosen based on two main properties: the usefulness of them for domain experts in the context of visualization and literature search, as well as the prevalence of those aspects within scientific abstracts. The latter point is crucial, since, while other aspects that only occur in the full-text might be important, this information is not available in our training pipeline, and would also not be available in common downstream settings like literature search. The aspects under consideration are the following: 

\begin{description}
    \item[The Hypothesis.] Research in invasion biology often focuses on overarching hypotheses that model the ecological relationships that explain why non-native species establish, spread, or impact ecosystems. As an example, the \textit{biotic resistance hypothesis} states that ecosystems with a higher species diversity are more resistant against invasions. For this factor, we experimented with distinguish between two granularities: the \textit{general} formulation of the hypothesis, which states the hypothesis in the most general and abstract form, and the \textit{specific} relationship under consideration, which contains more information on actual content of the study. Representing hypotheses at these two levels allows the model to connect studies addressing the same class of interactions while also capturing finer-scale mechanistic explanations. Within the revision process of the prompts, both formulations ended up being rather similar, though, thus leading only to minor differences in the summaries.
    \item[The ecosystem.] The characteristics of the recipient ecosystem strongly shape invasion outcomes. Relevant descriptors include climate regime (temperate, tropical, arid), habitat structure (e.g., freshwater stream, coastal saltmarsh, grassland), and ecological features that influence invasibility (such as disturbance, hydrological variability, or native species richness). Encoding the ecosystem context enables comparison of studies conducted in ecologically similar environments.  
    \item[The research question.] Scientific studies in invasion biology are often framed around central comparative questions, such as whether disturbance frequency and connectivity jointly determine invader success, or how predator abundance influences prey survival over time. Capturing the research question highlights the experimental or observational contrast being addressed, the variables of interest, and the scale of investigation. This aspect allows similarity comparisons to reflect whether studies are asking the same type of question, regardless of system or species.
    \item[The species] The traits and life-history of focal organisms are crucial for understanding invasion dynamics. Relevant descriptors include taxonomic level (plant, insect, fish, bird), growth or body form, trophic role (e.g., herbivore, predator), climate or biome affinity, dispersal and reproductive strategies, and invasion-relevant traits such as generalist diets or high reproductive output. By encoding functional characteristics rather than exact species identities, this aspect supports cross-study comparisons that generalize across taxa with similar ecological roles or invasion strategies.  
    \item[The methodology.] Research in invasion biology employs diverse approaches, from field surveys and long-term monitoring to manipulative experiments, laboratory trials, or modeling studies. Methodological differences influence both the type of evidence produced and the kinds of inferences that can be made. Encoding methodology ensures that similarity assessments account for whether studies used comparable designs, data sources, and analytical aims.  
    \item[The recommendations.] Many invasion biology studies conclude with explicit management or policy recommendations, such as prioritizing early detection, restoring native vegetation, or implementing targeted removal strategies. Capturing this information allows embedding-based comparisons to identify studies that converge not only in scientific findings but also in applied implications.
\end{description}  

\subsection{Summary Dataset Creation}
\label{app:inv_data_creation}

We use Mistral Small 3.1 \cite{mistral} to generate four summarizing sentences for each sample in the dataset and for each aspect under consideration. This model in particular was selected due to good performance in other (unpublished) experiments in the domain of invasion biology, thus making it the best candidate of comparable size at the time of summary generation.

To ensure good downstream performance of the resulting embedding models, we carefully design the prompts to enforce several key properties:  

\begin{enumerate}
    \item \textbf{Isolating aspects:} Prompts are structured so that the LLM focuses exclusively on the specified aspect, without introducing substantial information from other aspects. To achieve this, we explicitly instruct the model not to use species names or location names, encourage abstraction away from study-specific details, and provide illustrative examples that adhere to these constraints. We found that including such examples substantially improved the model’s adherence to the requirements.
    
    \item \textbf{Enforcing non-trivial matching:} To make the subsequent matching task during embedding model training non-trivial, prompts are designed to avoid overly specific information and instead focus on broader but relevant concepts. For example, species names are excluded in the \textit{species} aspect, geographic names in the \textit{ecosystem} aspect, and overly specific details in the \textit{hypothesis} aspect. Instead, the model is prompted to provide a broader semantic description of the underlying concept without naming it explicitly. This is crucial, since, during embedding model training, this prevents the model from encoding trivial information like species names for matching related summaries, and instead acquires a precise and generalizable semantic understanding of the underlying concepts.

    \item \textbf{Including similarity-relevant factors:} Prompts are designed to encourage summaries that incorporate a broad range of features relevant for assessing similarity between studies. This is essential, since only features that are consistently represented in the summarizing sentences can be learned as reliable indicators of similarity by the embedding models. For example, if summaries for the \textit{species} aspect were restricted to mentioning only species names, the model would learn to infer relatedness solely on the basis of identical names, treating all other species pairs with different names as equally unrelated. By contrast, including a richer set of features enables the embedding space to capture more nuanced relationships, positioning species with similar traits in close proximity. To this end, the LLM is explicitly instructed to include specific features whenever available. For the \textit{species} aspect, these are: "\textit{taxonomic level (plant, insect, fish, bird), growth or body form, trophic role (herbivore, omnivore, predator, detritivore), broad climate/biome affinity (temperate, tropical, boreal, arid), dispersal or reproductive mode (wind-dispersed seeds, planktonic larvae, broadcast spawner, clonal spread), and traits relevant to species invasions (e.g., generalist diet, high reproductive output, tolerance to salinity)}".

    \item \textbf{Handling non-applicable aspects:} The LLM is explicitly instructed to output \textit{“Not applicable.”} if the specific aspect does not apply to a study. This ensures that only highly informative sentences that teach the embedding model a precise understanding of the specific aspect are later included in the training process.
\end{enumerate}

An exemplary prompt for the \textit{hypothesis} aspect is shown in Figure \ref{fig:prompt_hypo}, while the full set of prompts is available in our GitHub repository.  

The final dataset sizes for each aspect are shown in Table \ref{tab:summary_dataset}. Note that only samples with at least two valid summaries are included, as only those can be used for embedding model training. Furthermore, the general and specific \textit{hypothesis} summaries are merged for training, thus enforcing that the embedding model learns to align generalized hypotheses with their more study-specific counterparts.  

\begin{table}[h!]
\centering
\begin{tabular}{lrr}
\toprule
\textbf{Aspect} & \textbf{Samples} & \textbf{Summaries} \\
\midrule
Hypothesis      & 37,709 & 300,889 \\
Ecosystem       & 35,684 & 140,480 \\
Research question & 37,718 & 150,778 \\
Species         & 36,695 & 145,919 \\
Methodology     & 37,389 & 148,795 \\
Recommendation  & 5,414  & 19,522  \\
\bottomrule
\end{tabular}
\caption{Dataset sizes for all aspects from the field of invasion biology.}
\label{tab:summary_dataset}
\end{table}

\begin{figure*}[h]
\centering
\begin{tcolorbox}[
    enhanced,
    colback=blue!8,
    colframe=blue!40,
    leftrule=4pt,
    rightrule=0pt,
    toprule=0pt,
    bottomrule=0pt,
    left=8pt, right=5pt, top=5pt, bottom=8pt,
    arc=0mm
]

\textbf{\footnotesize System Prompt:}
\vspace{2mm}

\footnotesize
You are a precise summary generator for scientific abstracts from the field of invasion biology. You will receive a scientific abstract from the field of invasion biology. For a single requested aspect, produce short, self-contained, declarative sentences that describe the aspect (as it is addressed in the scientific study) in general terms (no precise species names, no very specific place names, no numeric values, no references to "the study", "we", "this paper", or authors). If the requested aspect does not apply, return exactly: "Not applicable.". Return a strict JSON object as requested in the user instruction (no extra commentary).

\vspace{5mm}

\textbf{\footnotesize User Prompt:}
\vspace{2mm}

\footnotesize
This is the relevant scientific abstract:\vspace{0.5em}

\texttt{[ABSTRACT TEXT]}\vspace{0.5em}

Your task is the following:\vspace{0.5em}

Produce a single short declarative sentence that captures the broad, commonly-studied ecological relationship or directional hypothesis from the field of invasion biology that is addressed by the abstract. Use general causal or correlational language (e.g., "increases", "reduces", "facilitates", "is associated with") if applicable and avoid specific species names (general classes are okay), place names, numeric values, and any phrasing that references the paper or authors. Example: "Greater propagule pressure increases establishment probability across habitat types.".\vspace{0.5em}

Create a single comprehensive sentence that closely adheres to the requirements. Note that the given task might not apply to this study. In this case, simply reply "Not applicable.".\vspace{0.5em}

Return just the single sentence and nothing else. Create a self-contained sentence that makes sense without any additional context and summarizes the relevant factors without referencing "the study" or "the paper" (as done by the examples).\vspace{0.5em}

Use the following json response format: \{"sentence": "Your sentence."\}

\end{tcolorbox}

\caption{The prompt provided to Mistral Small 3.1 (24B) for generating summarizing sentences for the aspect \textit{Hypothesis (general)} within the field of Invasion Biology.}
\label{fig:prompt_hypo}
\end{figure*}

\begin{figure*}[htbp]
\centering

\begin{tcolorbox}[
  colback=gray!8,
  colframe=gray!65,
  boxrule=0.5pt,
  left=3mm,
  right=3mm,
  top=2mm,
  bottom=2mm,
  arc=1mm,
  enhanced,
  width=0.95\textwidth
]
\small
\textbf{Abstract title:} \emph{Over-invasion by functionally equivalent invasive species}

\vspace{1ex}
\textbf{Abstract:} Multiple invasive species have now established at most locations around the world, and the rate of new species invasions and records of new invasive species continue to grow. Multiple invasive species interact in complex and unpredictable ways, altering their invasion success and impacts on biodiversity. Incumbent invasive species can be replaced by functionally similar invading species through competitive processes; however the generalized circumstances leading to such competitive displacement have not been well investigated. The likelihood of competitive displacement is a function of the incumbent advantage of the resident invasive species and the propagule pressure of the colonizing invasive species. We modeled interactions between populations of two functionally similar invasive species and indicated the circumstances under which dominance can be through propagule pressure and incumbent advantage. Under certain circumstances, a normally subordinate species can be incumbent and reject a colonizing dominant species, or successfully colonize in competition with a dominant species during simultaneous invasion. Our theoretical results are supported by empirical studies of the invasion of islands by three invasive Rattus species. Competitive displacement is prominent in invasive rats and explains the replacement of R. exulans on islands subsequently invaded by European populations of R. rattus and R. norvegicus. These competition outcomes between invasive species can be found in a broad range of taxa and biomes, and are likely to become more common. Conservation management must consider that removing an incumbent invasive species may facilitate invasion by another invasive species. Under very restricted circumstances of dominant competitive ability but lesser impact, competitive displacement may provide a novel method of biological control.
\end{tcolorbox}

\vspace{0.6em}

\small  
\renewcommand{\arraystretch}{1.25}
\setlength{\tabcolsep}{8pt}

\begin{tabularx}{0.95\textwidth}{>{\raggedright\arraybackslash}X >{\raggedright\arraybackslash}X}
\toprule

\multicolumn{2}{c}{\textbf{Aspect: Hypothesis — General}} \\ \midrule
\textbf{LLM Summary:} Multiple invasive species can competitively displace one another, altering invasion success and biodiversity impacts.
&
\textbf{Reconstruction:} The presence of an established invasive species reduces the likelihood of successful invasion by a second invasive species. \\
\addlinespace[0.5em]

\multicolumn{2}{c}{\textbf{Aspect: Hypothesis — Specific}} \\ \midrule
\textbf{LLM Summary:} Invasive species may be replaced by other invasive species due to competitive interactions.
&
\textbf{Reconstruction:} Invasive species may outcompete established invasive species due to superior competitive abilities. \\
\addlinespace[0.5em]

\multicolumn{2}{c}{\textbf{Aspect: Ecosystem}} \\ \midrule
\textbf{LLM Summary:} Island ecosystems with a high degree of disturbance and frequent invasion events are considered.
&
\textbf{Reconstruction:} Island ecosystems with native rodent populations and frequent anthropogenic disturbance. \\
\addlinespace[0.5em]

\multicolumn{2}{c}{\textbf{Aspect: Research Question}} \\ \midrule
\textbf{LLM Summary:} How do competitive interactions between established and colonizing invasive species, mediated by relative advantage and propagule pressure, influence the displacement and coexistence of functionally similar invaders across various ecosystems?
&
\textbf{Reconstruction:} How does the competitive dynamics between two invasive species, one with an allelopathic effect and the other without, influence the establishment and spread of these species across different environmental conditions? \\
\addlinespace[0.5em]

\multicolumn{2}{c}{\textbf{Aspect: Species}} \\ \midrule
\textbf{LLM Summary:} Several small-bodied rodent species from temperate and tropical regions with omnivorous diet and high reproductive output.
&
\textbf{Reconstruction:} Small, generalist, omnivorous, invasive rodents from temperate regions with high reproductive output and tolerance to a wide range of habitats. \\
\addlinespace[0.5em]

\multicolumn{2}{c}{\textbf{Aspect: Methodology}} \\ \midrule
\textbf{LLM Summary:} A theoretical modeling framework explored population dynamics under various competitive scenarios to predict outcomes of invasive species interactions.
&
\textbf{Reconstruction:} A modeling approach with predictive simulations was used to explore interactions and outcomes between species with varying life history traits and competitive abilities. \\
\addlinespace[0.5em]

\multicolumn{2}{c}{\textbf{Aspect: Recommendation}} \\ \midrule
\textbf{LLM Summary:} Conservation management must consider that removing an incumbent invasive species may facilitate invasion by another invasive species.
&
\textbf{Reconstruction:} Managers should consider the relative fitness of native and invasive species and the potential for hybridization when choosing between eradication or coexistence management strategies. \\

\bottomrule
\end{tabularx}

\caption{Aspect-specific summaries and generated reconstructions for corresponding aspect-embeddings of SemCSE-Multi. The exemplary abstract by \cite{invbioabstract} is included in the test set.}
\label{fig:invasion_bio_example}
\end{figure*}

\subsection{Training Aspect Embedding Models}
\label{app:aspect_training}

To train the individual aspect-specific embedding models, we use SciDeBERTa \cite{10355927} as the base encoder (183M parameters), since it has been used by the original SemCSE paper and displayed strong results. Also using the same model eliminates a variable that could cause differences in results, so that using the same model enables a more precise relative performance assessment. The 768-dimensional output embeddings are projected to a 150-dimensional space using an additional linear projection head. For each aspect, training is performed only on samples with at least two valid summaries (i.e., summaries other than \textit{“Not applicable.”}). The corresponding dataset sizes are provided in Table \ref{tab:summary_dataset}.  

All models are optimized using AdamW \cite{loshchilov2019decoupledweightdecayregularization} with a learning rate of \(1\text{e}{-5}\), weight decay of \(1\text{e}{-4}\), and a batch size of 32. After every 1000 parameter updates, evaluation is conducted on a validation set consisting of 150 samples, each represented by two summaries for the respective aspect. During evaluation, the first summary of each sample is used as a query, while the set of second summaries across all samples forms the candidate pool. Performance is measured by the average rank at which the correct match is retrieved by the embedding model.  

\subsection{Training the Unified Model}
\label{app:unified_training}

For the unified model, we again use SciDeBERTa \cite{10355927} as the base encoder (183M parameters), augmented with one linear projection head per aspect. Each projection head maps the 768-dimensional \texttt{[CLS]} token embedding to a 150-dimensional aspect-specific representation.

To construct the training and validation datasets, we use the same samples as before and embed all summaries for all aspects using the corresponding aspect-specific embedding models. For each sample and aspect, we then compute a single ground-truth aspect embedding by averaging the individual summary embeddings for that aspect. The unified embedding model is then trained to predict these aspect embeddings directly from the full scientific abstract, using the appropriate projection head. To do this, we leverage an L2 loss, since it is the natural choice to facilitate precise reconstruction of known targets.

Training proceeds by sampling batches of 16 abstracts and computing the embedding reconstruction loss for all aspects applicable to each sample (i.e., aspects with at least one valid summary and thus a corresponding ground-truth embedding). Optimization is performed with AdamW using a learning rate of \(1\text{e}{-4}\) and weight decay of \(1\text{e}{-4}\).  

The optimal checkpoint is selected based on evaluation every 250 parameter updates. The validation set consists of 150 samples with ground-truth aspect embeddings, constructed in the same manner as for training. Evaluation is performed using the retrieval-based metric described in the previous section: for each aspect, the predicted embedding from SemCSE-Multi serves as the query, and the candidate pool consists of all ground-truth embeddings for that aspect in the validation set. The final evaluation score is reported as the average retrieval rank across all aspects.  

\subsection{LLM Assessment Creation}
\label{app:LLM_assessments}

For each aspect, we create similarity assessments between pairs of scientific abstracts that denote the similarity between the underlying studies with respect to just that specific aspect. This is done by leveraging an LLM, and the resulting scores are subsequently used to evaluate the correlation with the aspect-specific similarity rankings induced by our unified embedding model.

We selected 250 samples from the test dataset for each aspect (with fewer samples being used for the \textit{recommendation} aspect due to the low number of valid samples), ensuring that every selected abstract had at least one valid summary, which ensures that the aspect was indeed represented in the abstract. For every pair of abstracts in this subset, we prompted Qwen3-30B-A3B-Instruct-2507 \cite{yang2025qwen3technicalreport} to assign an integer similarity score, ranging from either 1–5 or 1–6 depending on the aspect, thereby quantifying the degree of similarity with respect to that aspect alone. 

Through extensive prompt design and experimentation, we identified several key components that were critical for achieving high-quality, aspect-specific similarity assessments:  
\begin{enumerate}
    \item \textbf{Explicitly specifying relevant and irrelevant information.} Listing the factors that should be considered when assessing similarity proved essential for guiding the model’s focus. Equally important was instructing the model to ignore irrelevant details such as specific species names, geographic locations, or overly study-specific information.
    \item \textbf{Providing a detailed scoring rubric.} A precise rubric outlining the requirements for each possible score was crucial for maintaining consistency across assessments.
    \item \textbf{Enforcing explicit reasoning.} Directly prompting the model to output an integer score yielded low-quality and inconsistent results. We found it essential to first instruct the model to summarize the relevant factors from both abstracts before assigning a similarity score. This mirrors the reasoning strategies employed by chain-of-thought models. In preliminary testing, we compared this condensed reasoning approach with the explicit reasoning by Qwen3-30B-A3B and found comparable accuracy. However, our approach was far more computationally efficient, making it feasible to process the large number of abstract pairs required for evaluation.
\end{enumerate}

An example prompt for the \textit{hypothesis} aspect is provided in Figure \ref{fig:prompt_pair}, with all other prompts available in our GitHub repository.  

\begin{figure*}[h!]
\centering
\begin{tcolorbox}[
    enhanced,
    colback=blue!8,
    colframe=blue!40,
    leftrule=4pt,
    rightrule=0pt,
    toprule=0pt,
    bottomrule=0pt,
    left=8pt, right=5pt, top=5pt, bottom=8pt,
    arc=0mm
]

\textbf{\footnotesize System Prompt:}
\vspace{2mm}

\footnotesize
You are a scientific research assistant from the field of invasion biology. You will be tasked with assessing the relatedness of two scientific studies from the field of invasion biology with regards to a specific aspect based on their abstracts. Note that both abstracts will for sure address the field of invasion biology, which in itself therefore shall not be treated as indicator for relatedness. Provide concise, accurate responses and adhere precisely to the information that is actually present in the abstracts.

\vspace{5mm}

\textbf{\footnotesize User Prompt:}
\vspace{2mm}

\footnotesize
Abstract A:\vspace{0.5em}

\texttt{[ABSTRACT 1]}\vspace{0.5em}

Abstract B:\vspace{0.5em}

\texttt{[ABSTRACT 2]}\vspace{0.5em}

Task: For the aspect 'general relationship / hypothesis', judge how similar the two abstracts are on the scale defined below.\vspace{0.5em}

These are the detailed instructions for this aspect: Compare the high-level ecological relationship or directional hypothesis addressed in the two abstracts. Abstract away from species, place names, numeric values, and study-specific outcomes. Focus on whether the same broad cause–effect relationship is being studied. Assess similarity based on:\vspace{0.5em}

(A) primary driver(s) / independent variable(s) (e.g., propagule pressure, disturbance, connectivity, enemy release, nutrient enrichment, climatic factor),

(B) primary response(s) / dependent outcome(s) (e.g., establishment probability, abundance, spread rate, impact on native diversity, survival, recruitment), and

(C) directional/causal framing or dominant mechanism (e.g., 'increases', 'reduces', 'facilitates', 'is associated with'; causal vs correlational framing; explicit mechanism like competition release or predation pressure).

(D) specific methods of measuring the high-level variables (e.g., how is "release from enemies" or "species diversity" quantified).\vspace{0.5em}

Important: Do NOT up-score similarity just because both papers are about biological invasions. This also means that many studies will measure "invasion success", which therefore shall not mean that they are related yet. For this specific variable, look at the more specific ways in which invasion success is measured/quantified, and at the remaining context of the high-level relationship. The default score is 1 unless there is a substantive match in some of the components. Minor overlaps (e.g., both about 'biotic interactions' without matching variables) should remain at 1 or 2.\vspace{0.5em}

Scoring rubric (use integers only):\vspace{0.5em}

\begin{etaremune}
\item Same high-level relationship with matching details — the primary driver(s) AND primary response(s) clearly match, AND the directional/causal framing or mechanism is the same or equivalent AND the high-level variables are quantified in strongly related ways.

\item Same high-level relationship without matching details — the primary driver(s) and primary response(s) clearly match (e.g., both evaluate if 'native species diversity decreases the likelihood of invasions'), but the scope is different and thus variables are quantified differently.

\item Related — a strong match on ONE core component (e.g., both mention species diversity, disturbance, enemy release, etc.) with some additional similar factor (e.g., same driver linked to a somewhat similar but not identical response, or same response but different classes of driver).

\item Weak relation — share only a minor conceptual element (e.g., both mention diversity, disturbance, or climate broadly) but the actual hypothesized driver–response link differs.

\item No meaningful overlap in the high-level relationship/hypothesis (beyond both being invasion studies and targeting species invasions).
\end{etaremune}

Return EXACTLY a single JSON object and nothing else with the following format:\vspace{0.5em}

\{"reasoning": <a short summary of the differences and similarities, and which category this supports>, "score": <integer score>\}\vspace{0.5em}

Do not include any other text, commentary, or examples.

\end{tcolorbox}

\caption{The prompt provided to Qwen3-30B-A3B-Instruct-2507 for creating pairwise similarity assessments between two scientific abstracts for the \textit{hypothesis} aspect.}\vspace{2em}
\label{fig:prompt_pair}
\end{figure*}

\subsection{Statistical Significance of Correlations}
\label{app:statistical_sig}

We evaluate the statistical significance of having higher correlation scores on the diagonals (i.e., between matching embedding and ground-truth aspects) compared to off-diagonal cells (i.e., cells that evaluate embeddings for one aspects against ground-truth for another). To this end, we perform a permutation-based significance test by creating the matrix $S_{a} \in \mathbb{R}^{n \times n}$ that contains the similarity scores between all pairs of $n$ scientific abstracts calculated using embeddings for aspect $a$, and the matrix $S_{b} \in \mathbb{R}^{n \times n}$ containing the same scores for embeddings of aspect $b$. We then repeatedly create a randomized matrix $S_r$ that contains, in each cell, either the score in the same cell of matrix $S_a$ or the score in the same cell of matrix $S_b$. We then evaluate the percentage of matrices $S_r$ that display a correlation with the ground truth of aspect $a$ that is at least as high as that for matrix $S_a$. For each possible pair of aspects and within 10,000 permutations, none of the permuted matrices showed a correlation as high as in the matching setting, thus demonstrating statistical significance. Exceptions are the (generally under-performing) \textit{recommendation} aspect, as well as the case with \textit{hypothesis} embeddings being evaluated against \textit{research question} ground-truth, since the \textit{hypothesis} embeddings already show stronger alignment than the matching embeddings even without randomization.

\subsection{A Unified Baseline}
\label{app:baseline}

SemCSE-Multi is based on a two-stage training procedure that first trains individual, aspect-specific embedding models using aspect-summaries and subsequently distills a unified model that predicts these embeddings directly from the abstract.

In this section, we evaluate an additional baseline that directly trains a unified model for all aspects at once in a single step. This is done by using the model structure of SemCSE-Multi (i.e., base encoder with one linear projection head for each aspect) and using the contrastive objective from Section \ref{sec:3}. For each parameter update, the contrastive objective is performed once for a separate batch for each aspect (and thus only applied to the specific linear projection head), thus updating the specific part of the model to gain a more precise understanding of that aspect.

Further, we now include complete abstracts in the objective, since otherwise the model would not learn to predict accurate embeddings from these longer and more information-rich texts. The task is therefore to match summaries of the same aspect to each other, and also to match abstracts to their aspect-specific summaries.

We find that the correlations between embedding-similarities and LLM-scores for matching aspects (i.e., the five on-diagonal scores, excluding the underperforming \textit{recommendation} aspect) are similar (42.6 for SemCSE-Multi and 42.8 for the baseline). However, the eighteen off-diagonal elements (excluding the cells matching \textit{hypothesis} to \textit{research question}, since these are sensible to be quite high) show a different picture: here, SemCSE-Multi achieves an average score of 21.43 compared to 28.3 for the baseline.

The reason for the reduced ability to disentangle the individual aspects is likely the joined process of training the baseline model for all aspects as once. While each aspect has its own embedding, these embeddings are all created by applying a linear projection to the same, shared, high-dimensional embedding produced by the encoder. For this reason, the update made to this embedding by one aspect can also affect the embedding of another aspect, and as long as this update does not reduce the matching performance on this other aspect, this change is not counteracted at any point. This ultimately leads to shared information between different aspects, and thus to reduced disentanglement.

Training each aspect in isolation instead ensures that only information useful for that aspect is included in the embedding, instead of also having irrelevant information from other aspects included.

\subsection{Training Embedding Decoders}

We decode embeddings back into natural language sentences by mapping embedding vectors into the input token space of an LLM and leveraging the LLM’s autoregressive language generation capabilities to create the natural language reconstruction.

To investigate the effect of model size and pre-training strategy, we evaluate two different LLMs: the purely autoregressive Llama 3 (8B) \cite{grattafiori2024llama3herdmodels} and the 24B-parameter chat model Mistral Small 3.1 \cite{mistral}. These models in particular were chosen due to, for the case of Mistral, being the model that generated the summaries in the first place, and, in the case on Llama, being simply a popular and smaller autoregressive LLM that has also been used in the original SemCSE paper \cite{brinner2025semcse}.  

We map the embedding into the input embedding space of the LLM using a dense neural network with a single hidden layer of dimension 768 and sigmoid activation. Each aspect embedding is projected into five sequential input-dependent tokens, which are accompanied by five additional trainable tokens that are input-independent and are prepended to the input-dependent tokens to provide a general task description to the LLM, independent of the specific embedding. For Llama, the resulting input sequence consists of the \texttt{<bos>} token followed by the ten trainable/decoded tokens, plus one additional trainable token acting as delimiter to the generated text. For Mistral, all 10 trainable/decoded tokens are inserted into the user prompt while preserving the standard chat format, such that the generated response by the LLM is produced as standard assistant response.  

Because of the different hidden dimensionalities of the two LLMs, the transformation networks contain approximately 16M parameters for Llama and 24M parameters for Mistral. During training, only the parameters of these mapping networks are updated, while the LLM weights remain frozen.  

Training is performed using the dataset of summarizing sentences described in Section \ref{app:inv_data_creation}. For each sample and aspect, we compute the average of the aspect embeddings across all available summaries to obtain a single aspect embedding for that sample. The model is then trained to reconstruct all of the corresponding summarizing sentences, with one of them being selected at random each time that sample is used. The trainable parameters are optimized using a standard language modeling loss to reconstruct the original summary from the embedding.

Optimization is carried out with AdamW, using a batch size of 4, a learning rate of \(1\text{e}{-3}\), and weight decay of \(1\text{e}{-4}\). Model checkpoints are evaluated every 250 parameter updates on a validation set of 150 samples, and the best-performing checkpoint is selected based on reconstruction loss.

\subsection{t-SNE Details}
\label{app:tsne}

t-SNE \cite{vandermaaten08a} is a widely used dimensionality reduction technique due to its favorable balance between effectiveness and conceptual simplicity. Given a pairwise distance matrix $A \in \mathbb{R}^{n \times n}$ for $n$ samples in the original embedding space, t-SNE learns a low-dimensional representation $\{y_1, \dots, y_n\}$ with $y_i \in \mathbb{R}^d$ (typically $d=2$ or $3$), such that the local similarity structure of the data is preserved.  

In the high-dimensional space, conditional probabilities are defined as  
\[
p_{j|i} = \frac{\exp\!\left(-A_{ij}^2 / 2\sigma_i^2\right)}{\sum_{k \neq i} \exp\!\left(-A_{ik}^2 / 2\sigma_i^2\right)}, 
\quad p_{ii} = 0,
\]  
which are then symmetrized as  
\[
p_{ij} = \frac{p_{i|j} + p_{j|i}}{2n}
\]  

In the low-dimensional embedding, similarities are modeled using a heavy-tailed Student-$t$ distribution with one degree of freedom:  
\[
q_{ij} = \frac{\left(1 + \|y_i - y_j\|^2\right)^{-1}}{\sum_{k \neq l} \left(1 + \|y_k - y_l\|^2\right)^{-1}}, 
\quad q_{ii} = 0.
\]

The embedding $Y = [y_1, \dots, y_n]^\top \in \mathbb{R}^{n \times d}$ is then optimized by minimizing the Kullback–Leibler divergence between the distributions $P = (p_{ij})$ and $Q = (q_{ij})$:  
\begin{align}
\mathcal{L}(Y) = \mathrm{KL}(P \,\|\, Q) = \sum_{i \neq j} p_{ij} \log \frac{p_{ij}}{q_{ij}}
\end{align}

This objective encourages points that are close in the original space to remain neighbors in the low-dimensional representation, while the heavy-tailed distribution prevents unrelated points from collapsing together.

\subsection{Details on Interactive Embedding and Decoding}
\label{app:interactive_details}

Section \ref{sec:6} introduces approaches for interacting with the low-dimensional representation of our semantic embedding space. For visualization, we employ t-SNE and compute pairwise distances between samples based on their embeddings using $d(e_i, e_j) = 1 - \cos(e_i, e_j)$ as distance between two embeddings \(e_i\) and \(e_j\), with $\cos$ denoting the cosine similarity.

\paragraph{Embedding additional samples.}  
To embed a new sample $x_{n+1}$ into an existing visualization, we first compute its high-dimensional distances to all existing samples:
\[
A_{(n+1)j} = d(x_{n+1}, x_j), \quad j = 1, \dots, n.
\]
These distances are used to form conditional probabilities $p_{j|n+1}$ and $p_{n+1|j}$ in the same way as for the original samples, followed by symmetrization to obtain $p_{n+1,j}$.

We then fix all existing low-dimensional coordinates $\{y_1, \dots, y_n\}$ and optimize only the new coordinate $y_{n+1}$ such that the KL divergence objective in Equation~\ref{eq:tsne_optim} is minimized:
\[
\min_{y_{n+1}} \; \mathcal{L}(Y \cup \{y_{n+1}\}) 
= \sum_{i \neq j} p_{ij} \log \frac{p_{ij}}{q_{ij}},
\]
where $q_{ij}$ is recomputed with $y_{n+1}$ included. This ensures that the new point is positioned in the low-dimensional space such that its local neighborhood structure reflects the distances in the original embedding space, while the configuration of existing points remains unchanged.

To speed up convergence, we initialize $y_{n+1}$ as the mean of the low-dimensional embeddings corresponding to the five nearest neighbors within the original, high-dimensional embedding space.

\begin{table*}[ht!]
\renewcommand{\arraystretch}{1.00}
\centering
\resizebox{0.6\textwidth}{!}{
\begin{tabular}{lcccccc}
\toprule
 & Disease & Methodology & Patient Group \\
\midrule
Samples & 282 & 396 & 330 \\
MRR \footnotesize Main model & .859 & .741 & .694 \\
MRR \footnotesize Other models & .778 & .541 & .546 \\
MRR \footnotesize SemCSE & .811 & .548 & .510 \\
\bottomrule
\end{tabular}}
\caption{Retrieval performance measured via Mean Reciprocal Rank (MRR \%), multiplied by 100. 
Main = model trained for the aspect, Others = average of non-matching models.}
\label{tab:medical_individual_retrieval}
\end{table*}

\paragraph{Reconstructing high-dimensional embeddings.}  
Conversely, when a user specifies a new point $y_{n+1}$ in the visualization, we seek to infer a high-dimensional embedding $e_{n+1}$ that would plausibly map to this position. To this end, we keep all existing high-dimensional embeddings $\{e_1, \dots, e_n\}$ fixed and introduce $e_{n+1}$ as an optimizable variable.

The distances in the high-dimensional space are defined as
\[
A_{(n+1)j} = d(e_{n+1}, e_j), \quad j = 1, \dots, n,
\]
which yield new conditional probabilities $p_{n+1,j}$. In the low-dimensional space, the user-specified point $y_{n+1}$ is treated as fixed. We then minimize the same KL divergence objective:
\[
\min_{e_{n+1}} \; \mathcal{L}(Y \cup \{y_{n+1}\}, E \cup \{e_{n+1}\}),
\]
with respect to $e_{n+1}$ only.

In practice, we constrain $e_{n+1}$ to the top $k=20$ principal components of the original embedding space, i.e.,
\[
e_{n+1} = U z, \quad z \in \mathbb{R}^k,
\]
where $U \in \mathbb{R}^{d \times k}$ are the leading principal components. This reduces the optimization problem to a low-dimensional search over $z$ while ensuring that the reconstructed embedding remains within the most semantically meaningful subspace.

Further, we initialize the embedding $e_{n+1}$ as the mean of the high-dimensional embeddings corresponding to the five nearest neighbors within the low-dimensional visualization to speed up convergence.

Once optimized, the resulting $e_{n+1}$ can be decoded via the pipeline introduced in Section~\ref{sec:5}, providing a natural language explanation of the semantic content represented by the chosen location in the visualization.

\subsection{Details on Embedding Reconstruction Decoding}
\label{app:reconstruc_experiment}

We evaluate our framework’s ability to generate semantic descriptions for previously unoccupied positions in low-dimensional visualizations, which is performed separately for each aspect. For a given aspect, we select all test samples with at least one valid summary, compute their aspect-specific embeddings with SemCSE-Multi, and construct a low-dimensional embedding space using t-SNE.  

In the t-SNE calculation, each time one sample is omitted and serves as the target for reconstruction, which is repeated so that every sample in the dataset is left out once, and the final score is obtained by averaging across all iterations.

After computing the low-dimensional embeddings, the omitted sample’s high-dimensional embedding $e_{n+1}$ is temporarily added to extend the distance matrix $A$. The corresponding low-dimensional coordinate $y_{n+1}$ is then inferred using the embedding procedure described in Appendix \ref{app:interactive_details}. We subsequently remove $e_{n+1}$ from the high-dimensional set and optimize a new high-dimensional embedding $\tilde{e}_{n+1}$ for the position $y_{n+1}$, again following the method in Appendix \ref{app:interactive_details}.  

Finally, the decoding procedure from Section \ref{sec:5} is applied to $\tilde{e}_{n+1}$ with the goal of reconstructing the aspect-specific summaries of the left-out sample. Perplexity is computed for each summary and averaged to yield a single score per sample, and the overall evaluation metric is obtained by averaging these scores across all samples.

\section{Experiments on the Medical Domain}
\label{sec:medical_appendix}

\subsection{Introduction}

To establish wider applicability of our proposed multifaceted embedding framework beyond the domain of invasion biology, we perform an additional evaluation in the medical domain. Due to the computational resources required for the model training and generation of training and test data, we limit this second evaluation to a smaller scale, thus only including three factors and limiting the training set to 15,000 samples and the number of samples for pairwise similarity assessment to 200.

\begin{table*}[t]
\libertine
\footnotesize
\setlength{\tabcolsep}{14pt} 
\renewcommand{\arraystretch}{1.55}
\libertineSB{
\begin{minipage}[t]{0.47\textwidth}
\resizebox{\textwidth}{!}{
\begin{tabular}{clccccccc}
\multicolumn{2}{r}{\raisebox{0em}[0pt][0pt]{\makebox[0pt][r]{\shortstack[r]{Ground Truth\\Similarities $\rightarrow$}}}} \hspace{-25px} & \diaghead{{Disease}} & \diaghead{{Methodology}} & \diaghead{{Patient Group}} \\
\midrule
\multirow{4}{*}{\hspace{-10px}\rotatebox{90}{\underline{{SemCSE-Multi}}} \hspace{-30px}}
  & \hspace{-20px}{Disease} & \HeatCell{20}{20.29} & \HeatCell{7}{7.57} & \HeatCell{7}{7.81} \\
  & \hspace{-20px}{Methodology} & \HeatCell{10}{10.97} & \HeatCell{38}{38.40} & \HeatCell{18}{18.66} \\
  & \hspace{-20px}{Patient Group} & \HeatCell{12}{12.18} & \HeatCell{28}{28.45} & \HeatCell{41}{41.62} \\
  & \hspace{-20px}{Combined} & {20.29} & {38.40} & {41.62} \\ \midrule
  \multirow{6}{*}{\hspace{-10px}\rotatebox{90}{\underline{{Baselines}}} \hspace{-30px}}
  & \hspace{-20px}{SPECTER}  & {31.16} & {25.28} & {20.36} \\
& \hspace{-20px}{SPECTER2} & {30.24} & {18.70} & {17.50} \\
& \hspace{-20px}{SciNCL}   & {32.75} & {27.75} & {18.25} \\
& \hspace{-20px}{SemCSE}   & {29.14} & {33.30} & {28.33} \\
& \hspace{-20px}{MedEmbed-base}   & {30.98} & {31.78} & {18.56} \\
& \hspace{-20px}\resizebox{8em}{!}{Clinical-ModernBERT} & {14.81} & {39.94} & {25.03} \\
\bottomrule
\end{tabular}}

\caption{Correlation matrix showing Spearman correlations between pairwise similarities computed from embedding models (aspect embeddings from SemCSE-Multi as well as baselines) and the LLM-generated pairwise ground-truth similarities.}

\label{tab:medical_predictor_label_heatmap}
\end{minipage}
\hspace{1em}
\begin{minipage}[t]{0.47\textwidth}
\resizebox{\textwidth}{!}{
\begin{tabular}{clccccccc}
\multicolumn{2}{r}{\raisebox{0em}[0pt][0pt]{\makebox[0pt][r]{\shortstack[r]{Ground Truth\\Similarities $\rightarrow$}}}} \hspace{-25px} & \diaghead{{Disease}} & \diaghead{{Methodology}} & \diaghead{{Patient Group}} \\
\midrule
\multirow{4}{*}{\hspace{-10px}\rotatebox{90}{\underline{{SemCSE-Multi}}} \hspace{-30px}}
  & \hspace{-20px}{Disease} & \HeatCell{36}{50.35} & \HeatCell{15}{21.90} & \HeatCell{29}{41.15} \\
  & \hspace{-20px}{Methodology} & \HeatCell{13.3}{19.40} & \HeatCell{38}{54.05} & \HeatCell{31}{45.95} \\
  & \hspace{-20px}{Patient Group} & \HeatCell{8}{23.70} & \HeatCell{24}{34.25} & \HeatCell{44}{62.65} \\
  & \hspace{-20px}{Combined} & {50.35} & {54.05} & {62.65} \\ \midrule
  \multirow{6}{*}{\hspace{-10px}\rotatebox{90}{\underline{{Baselines}}} \hspace{-30px}}
  & \hspace{-20px}{SPECTER}  & {41.10} & {32.35} & {46.00} \\
& \hspace{-20px}{SPECTER2} & {38.45} & {27.70} & {40.90} \\
& \hspace{-20px}{SciNCL}   & {43.50} & {31.05} & {45.75} \\
& \hspace{-20px}{SemCSE}   & {40.00} & {37.20} & {46.75} \\
& \hspace{-20px}{MedEmbed-base}   & {42.30} & {31.60} & {46.95} \\
& \hspace{-20px}\resizebox{8em}{!}{Clinical-ModernBERT} & {25.00}   & {41.85} & {47.50} \\
\bottomrule
\end{tabular}}

\caption{Top-10 retrieval performance using pairwise similarities computed from embedding models (aspect embeddings from SemCSE-Multi as well as baselines). Metric: Percentage of samples in the top-10 retrieved items that are also within the top-10 most similar items according to the LLM assessments.}
\label{tab:medical_predictor_label_heatmap_top10}
\end{minipage}}
\hfill
\end{table*}

\subsection{Dataset Creation and Aspect Definition}

For creating the dataset, we sampled a random set of 15,000 scientific abstracts from the publicly available PubMed database. The first 500 samples thereby serve as test samples, while the subsequent 150 samples are used as validation samples in the individual training runs. Both the summary generation for all splits as well as the pairwise similarity assessment generation for 200 test samples with regards to the individual aspects are done in the same experimental setting as for the domain of invasion biology. The adapted prompts are again available in our GitHub repository.

Unlike for the domain of invasion biology, we were not consulted by domain experts for the experiments in the medical domain. For this reason, the decision on suitable aspects for the multifaceted embedding model was done with the help of popular Chat LLMs, with a focus on selecting aspects that are both relevant for the domain, but at the same time might not be necessarily captured well by existing embedding models, to underscore the great adaptability to otherwise neglected aspects.

We settled on the following aspects:
\begin{description}
    \item[The disease:] Many scientific studies from the medical domain center around a specific disease or medical condition. We prompted the LLM to focus on general factors that would be useful to assess similarity between different diseases (e.g., disease class, dominant pathological process, or hallmark clinical features).
    \item[The methodology:] The specific methodological approach underlying a scientific study in the medical domain is of high importance since it yields crucial information about the inferences that can be drawn from the reported results. The LLM was prompted to summarize information about the study design class (e.g., randomized controlled trial, prospective cohort study), the data source and collection method (e.g., clinical measures, imaging, electronic health records) and the overall analytic approach (e.g., longitudinal follow-up, comparative analysis).
    \item[The patient group:] The patient group under consideration within a given study is a key factor for assessing the applicability to a specific patient in a clinical setting. For this reason, we prompted the LLM to describe the patient group, including factors like "age group, sex when relevant, general health status or risk profile, severity or stage descriptors, comorbidity burden, or whether the subjects are experimental animals".
\end{description}

We hypothesize that this aspect selection provides valuable insights into the relative strength of our multifaceted embedding model compared to standard, mostly citation-based embedding models, since these likely perform well with regards to assigning high similarity to studies addressing the same medical condition (since such studies likely cite each other with high probability), while scoring lower with regards to the methodology and patient group.

\subsection{Individual Embedding Model Evaluation}

The training of the individual embedding models is done exactly like for the experiments in the domain of invasion biology. The evaluation results for the summary-retrieval setting described in Section \ref{sec:individual_model_evaluation_setup} are displayed in Table \ref{tab:medical_individual_retrieval}.

Similar to the experiments on the domain of invasion biology, we see the highest retrieval performance if the specialized aspect-specific model is used, while retrieval performance using the embedding models for the non-matching aspects and the SemCSE baseline is noticeably lower. Importantly, we observe a generally very high matching performance, but note that the higher performance compared to the invasion biology experiments is caused by the lower dataset size. Nevertheless, the high scores indicate that the correct match is usually retrieved as first or second candidate.

\begin{table*}[ht]
\setlength{\tabcolsep}{2pt} 
\centering
\begin{tabular}{lccccccc}
\toprule
& \multicolumn{4}{c}{Llama} & \multicolumn{3}{c}{Mistral} \\
\cmidrule(lr){2-5} \cmidrule(lr){6-8}
Aspect & Matching & Reconstructed & Shuffled & Unconditioned & Matching & Reconstructed & Shuffled \\
\midrule
Disease & 6.28 & 11.76 & 25.20 & 30.08 & 6.44 & 12.75 & 26.78 \\
Methodology & 6.71 & 10.63 & 26.73 & 167.37 & 6.91 & 10.90 & 30.14 \\
Patient Group & 7.39 & 11.02 & 39.44 & 111.86 & 7.49 & 11.46 & 48.73 \\
\bottomrule
\end{tabular}
\caption{Perplexity results for Llama and Mistral decoders under normal and shuffled (ablation) settings. Lower perplexity indicates better performance.}
\label{tab:medical_decoder_perplexity}
\end{table*}

\subsection{Unified Model Evaluation}

The training of the unified SemCSE-Multi embedding model is done exactly like for the experiments in the domain of invasion biology. The evaluation results for the correlation-based evaluation described in Section \ref{sec:unified_evaluation_setup} are displayed in Table \ref{tab:medical_predictor_label_heatmap}.

Note that for the medical domain, due to the availability of specialized, domain-specific embedding models, we added two additional baseline models in the form of MedEmbed-base-v0.1 \cite{balachandran2024medembed} and Clinical-ModernBERT \cite{lee2025clinical} to the evaluation.

Again, we observe the highest correlation between aspect-specific embeddings and pairwise aspect evaluations on the diagonal entries, thus indicating that the aspect-specific embeddings accurately represent and isolate these aspect-specific similarities.

Especially for the two aspects that we hypothesized to be underrepresented by the general embedding models (i.e., \textit{methodology} and \textit{patient group}), no baseline general-domain model can rival the performance of our multi-faceted embedding model, with only Clinical-ModernBERT surpassing our model slightly in the \textit{methodology} aspect while falling far short on the other two aspects.

In contrast, for the \textit{disease} aspect, many baseline models outperform our multifaceted embedding approach. We attribute this primarily to suboptimal prompt design for the generation of training summaries. Specifically, the prompts did not sufficiently instruct the LLM to extract features that enable meaningful similarity assessments between non-identical diseases - for instance, shared pathophysiological mechanisms or affected organ systems. As a result, the model likely learned an embedding space that effectively clusters closely related diseases but fails to capture nuanced similarities among less directly related ones.

To test this hypothesis, we conducted a follow-up experiment using the same LLM-based pairwise similarity assessments. While our initial evaluation measured the overall correlation between each sample’s embedding-induced and LLM-assessed similarity rankings across the full dataset, we now instead focus on the precision of the top retrieved items. Concretely, we assess the proportion of the ten most similar samples (according to the embedding model) that also appear among the ten most similar samples identified by the LLM.

The results of this modified evaluation, shown in Table \ref{tab:medical_predictor_label_heatmap_top10}, reveal a substantial improvement of our multifaceted embedding model on the disease aspect. It now surpasses all baselines by far, demonstrating its superior ability to retrieve papers concerning closely related diseases. The lower performance in the global correlation analysis can therefore be attributed to its limited ability to order less related diseases correctly, rather than to deficiencies in identifying truly similar ones.

These findings underscore the critical importance of well-designed prompts. Achieving optimal performance likely requires the careful formulation of aspect-specific prompts, ideally guided by domain experts to ensure that the LLM captures the most relevant conceptual relationships.

\subsection{Embedding Decoder Evaluation}

We train the embedding decoder equivalently to the experiments in the domain of invasion biology. The evaluation results for both the normal and reconstructed decoding described in Sections \ref{sec:decoder_eval_setup} and \ref{sec:t_SNE_evaluation_setup} are displayed in Table \ref{tab:medical_decoder_perplexity}.

Consistent with the findings in invasion biology, the decoder achieves low perplexity scores when reconstructing the high-dimensional embeddings, substantially outperforming the \textit{shuffled} and \textit{unconditioned} baselines that reconstruct either non-matching summaries from the same aspect or evaluates perplexity of the summaries without any conditioning. This confirms that the semantic information encoded in the aspect-specific embeddings are accurately translated back into natural language, thus effectively creating human-interpretable descriptions of the embedding’s content.

As expected, the \textit{reconstructed} embeddings yield slightly higher perplexity values than the direct decoding of original embeddings. Nevertheless, the results clearly demonstrate that meaningful and coherent sentences can be generated even for embeddings optimized for points in low-dimensional visualizations, thereby validating the feasibility of making user-specified regions in such visualizations interpretable.

\end{document}